\documentclass[preprint, nonacm]{jair}
\usepackage{dirtree}
\usepackage{microtype}
\usepackage{subcaption}
\usepackage{footmisc}

\setcopyright{cc}
\acmDOI{10.1613/jair.1.xxxxx}

\JAIRAE{Christian Bessiere}
\JAIRTrack{} % Insert JAIR Track Name only if part of a special track
\acmVolume{0}
\acmArticle{0}
\acmMonth{10}
\acmYear{2026}

\RequirePackage[
  datamodel=acmdatamodel,
  style=acmauthoryear,
  backend=biber,
  giveninits=true,
  uniquename=init
  ]{biblatex}

\begin{document}

%\title{JAIR Example Template}
\title[MPMMine]{MPMMine: Better Benchmarks for Constraint Acquisition}

%% used to denote shared contribution to the research and/or corresponding author.
\author{Rafał Stachowiak}
\orcid{0009-0002-7978-900X}
\email{rstachowiak@cs.put.poznan.pl}
\affiliation{%
  \institution{Poznan University of Technology}
  \city{Poznan}
  \country{Poland}
}

\author{Tomasz P. Pawlak}
\authornote{Corresponding Author.}
\orcid{0000-0002-8353-0562}
\email{tpawlak@cs.put.poznan.pl}
\affiliation{%
  \institution{Poznan University of Technology}
  \city{Poznan}
  \country{Poland}
}

%% The short list of authors must be made of the list of all authors' lastnames.
\renewcommand{\shortauthors}{Stachowiak \& Pawlak}
%% If this is too long and overlaps other information printed in the page headers, use
%\renewcommand{\shortauthors}{Xu et al.}

\begin{abstract}
    {\bf Background:} 
    Constraint Acquisition (CA) and related research on the validation and enhancement of Mathematical Programming (MP) models using domain knowledge artifacts are currently limited by inadequate benchmarks. The existing benchmarks are loosely organized, provide inconsistent types and formats of artifacts for individual problems, are designed for benchmarking solvers rather than CA algorithms, include narrow classes of problems, %, precluding problem-wide experiments. Many datasets are crafted for benchmarking solvers rather than CA algorithms. %, providing data at a level of abstraction that is not compatible with the requirements of CA methods. 
    or omit crucial details for the reproducibility of experiments and fair comparison of CA methods.
    
    {\bf Objectives:}
    We develop a benchmark dataset, a supporting software package, and a benchmarking tool crafted for the consistent, systematic, and comprehensive assessment of CA algorithms, bridging the gap in algorithmic development and experimental infrastructure.
    
    {\bf Methods:}
    This work presents MPMMine, a benchmark suite designed to assess algorithms that discover, validate, and enhance MP models using diverse artifacts. The development of MPMMine is guided by design rules supporting consistency, standardization, completeness, extensibility, openness, and version control. It adopts a uniform structure and relies on open formats. %: MiniZinc, CommonMark, and JSON. 
    It provides alternative formulations of MP models per problem, tens of parameter instantiations per MP model, and thousands of exemplary solutions and non-solutions in discrete, continuous, and mixed domains, alongside natural-language descriptions to support text-to-model methods. The mpmmine-py Python package provides an application programming interface for accessing the dataset artifacts. The MPMMine Leaderboard is a~tool that systematically evaluates existing CA algorithms using MPMMine under consistent conditions. It leverages open source implementations, runs them in isolated containers adapting input/output to their needs to build an extensive, interactive scoreboard.
    
    {\bf Results:}
    MPMMine consists of 20 operations research problems from various real-world problem domains, %, including scheduling, cutting, packing, routing, locating, blending, and timetabling
     with 28 MP models and 387 instances in total. % for diverse variants and parameterizations. 
    MPMMine provides ca.~six million unique artifacts for training and testing MP models.
    The MPMMine Leaderboard shows a systematic comparison of four state-of-the-art CA algorithms.
    
    {\bf Conclusions:}
    An MP model encoded using a high-level modeling language provides extra information about relationships between parameters and variables that are concealed in low-level models, making it a rich base for the verification of certain features of the CA algorithm.
    %Constraint acquisition and model solving benefit from different MP model encodings of the same computational problem. The representational bias of training data, e.g., loose variable values vs arrays, influences the encoding that the MP model can take. 
    Different MP models for the same problem enable the verification of the performance of CA algorithms under varying problem encodings. Parameterized MP models and multiple text descriptions enable the assessment of how algorithm performance scales with the volume and variety of training data. %While larger training datasets usually lead to better fitting, building MP models applicable to different parameterizations turns out challenging.
    The distribution of exemplary solutions and non-solutions influences algorithmic performance, as sample concentrations may unduly contribute to acquiring specific relationships while neglecting others. Distribution shaping enables verification of varying learning scenarios, but controlling the distribution by drawing directly from the highly dimensional and highly constrained solution space of an MP model is technically challenging. A~large uniform precomputed sample enables simplified distribution shaping through, e.g., oversampling. Large samples lead to significant dataset bloat. Bundling the dataset into a single compressed file reduces latency several times compared to direct file system access and reduces disk space usage by 5-6 times. The mpmmine-py package offers transparent access to individual artifacts independent of the actual dataset representation. 

\end{abstract}

%% JAIR Note: 
%% Do not include ACM CCS Concepts or Keywords

%% To be updated by authors.
\received{9 October 2026}
%\received[accepted]{X XXX 2026}

\maketitle

\section{Introduction}\label{sec:intro}

\subsection{Background}\label{sec:intro:back}
A Mathematical Programming (MP) model \cite{williams2013model} is a~formalism for expressing computational problems. It consists of variables, which represent the unknown quantities to be determined, constraints that capture the relationships among them, and an objective function that evaluates candidate solutions. High-level MP languages, such as MiniZinc~\cite{minizinc1}, further extend this framework by supporting parameters, data structures, and succinct constructs based on quantifiers, set algebra, and global constraints that replace long lists of low-level equations.

The manual construction of MP models is labor-intensive and demands expertise in both the application domain and the underlying optimization framework. Even for skilled practitioners, designing a high-quality MP model for realistic instances with thousands of variables and constraints can often take weeks. \emph{Constraint Acquisition} (CA) (e.g., \citet{bessiere2017constraint,10.1613/jair.1.14752,Menguy2025,10.1145/3205455.3205480,Pawlak2021,Tsouros2020,Mechqrane2026,PAWLAK2021100896}) addresses this challenge by providing algorithms that discover MP models from domain knowledge artifacts, chiefly from example solutions or answers to queries about the classification of example solutions. CA thus supports experts in deriving MP models from existing data. At the same time, CA naturally fits into a broader setting in which one not only discovers new MP models, but also checks conformance and enhances existing models using domain knowledge artifacts. To differentiate CA from this broader view, we introduce the umbrella term \emph{MP model mining} (MPMM), which covers all tasks concerned with the discovery, conformance checking, and enhancement of MP models based on domain knowledge.

The existing literature on MPMM is substantial---see surveys by \citet{STACHOWIAK2026100905,Cappart2025}, reflecting active research. Nonetheless, that recent survey papers reveal that the majority of contributions remain at the level of basic research and do not progress beyond Technology Readiness Level (TRL) 3. Reaching laboratory validation (TRL 4–5) through extensive testing in simulated environments remains difficult since basic MPMM research is hindered by several issues that slow or prevent further advancement, e.g.:
\begin{enumerate}
    \item \emph{Non-standard benchmarks} -- researchers spend significant effort on data acquisition and curation; the evaluation sets of computational problems appear in multiple variants, each admitting different MP encodings and instances; this often leaves open questions like, e.g., Is the chosen problem encoding tailored to favor the algorithm under evaluation? On what basis was a particular benchmark set selected?
    \item \emph{Non-reproducible experiments} -- experimental protocols frequently omit crucial details, and open questions remain, e.g., How were the training and test solutions generated? What is the distribution of the instances? What bounds on constraint cardinality were used?
    \item \emph{Incomparable results} -- the use of custom experimental setups leads to results that are difficult to compare; e.g., Are reported performance differences between papers attributable to genuine algorithmic progress or to differing experimental methodologies?
    \item \emph{Limited instances} -- studies evaluating a single problem instance fail to reflect the true performance of an algorithm on the entire problem class; e.g., How does the algorithm scale as the problem instance size increases?
    \item \emph{Scarcity of continuous and mixed problems} -- problems with continuous variables are often overlooked in MPMM research, despite their practical relevance and commonness in Operations Research (OR); e.g., How were inexact variable values or parameters treated? How were tolerance levels set?
\end{enumerate}
We argue that these shared difficulties stem mainly from the absence of dedicated research infrastructure for MPMM, as advocated by \citet{STACHOWIAK2026100905,Cappart2025}. Existing benchmark suites, such as CSPLib \cite{csplib}, MiniZinc challenge \cite{minizinc2}, and MIPLIB \cite{MIPLIB}, were created to evaluate solvers rather than MPMM methods. Moreover, these benchmarks lack strict standardization: instances differ in file organization, use heterogeneous modeling languages, e.g., MiniZinc \cite{minizinc1}, OPL \cite{OPL}, Essence~\cite{essence}, CPMpy \cite{cpmpy}, MPS \cite{gurobi}, and often omit or inconsistently store domain knowledge artifact, e.g., solutions, parameters, descriptions. As a~result, MPMM studies typically employ ad-hoc experimental setups, in which missing artifacts, such as exemplary solutions, are generated via custom procedures that are rarely documented.

\subsection{Contributions}\label{sec:intro:contrib}

The contributions of this work are threefold.

First, we present \emph{MPMMine}\footnote{\label{fn:mpmmine}\url{https://github.com/mpmmine/mpmmine}} benchmark dataset designed to evaluate algorithmic performance across the full spectrum of MPMM problems: discovery, conformance checking, and enhancement. MPMMine development is guided by six key principles---\emph{consistency}, \emph{standardization}, \emph{completeness}, \emph{extensibility}, \emph{openness}, and \emph{version control}---which are detailed in Section~\ref{sec:mpmm:rules}. MPMMine currently comprises 20 problems, including combinatorial, continuous, and mixed types, enriched with diverse domain knowledge artifacts, and it continues to grow, with each problem associated with multiple instances, natural-language descriptions, and thousands of solutions and non-solutions per instance. All data were generated via a~well-documented protocol, facilitating experimental reproducibility and enabling fair comparisons of different studies through shared training/test sets. The benchmark format is standardized, so larger collections of problems can be used without modifying existing code. For each MP model, instances span a~range of sizes and complexities, aiding the evaluation of scalability. Every model is paired with a model checker confirming feasibility of solutions and infeasibility of non-solutions. 

Second, we introduce \emph{mpmmine-py}\footnote{\url{https://github.com/mpmmine/mpmmine-py}} Python package. % that facilitates the use of MPMMine. % from Python code. 
This makes the use of MPMMine straightforward in the most common programming language in artificial intelligence research. What is more, \emph{mpmmine-py} is highly optimized, and together with the compressed package of MPMMine, it provides significantly smaller access latency to individual artifacts than direct file system access (though specific speedups are hardware-dependent).

Third, we evaluate various MPMM discovery algorithms using MPMMine under the \emph{MPMMine Leaderboard} initiative \cite{MPMMineLeaderboard}.\footnote{\label{fn:leaderboard}\url{https://github.com/mpmmine/leaderboard}} This enables us to create an~interactive scoreboard and fairly compare MPMM algorithms against each other. The detailed comparison protocol is described in Section~\ref{sec:leaderboard}.

\section{Related work}\label{sec:related}

Almost all existing benchmark collections of MP models were developed for evaluating solvers rather than algorithms that mine MP models using domain knowledge artifacts. Here, we highlight the deficiencies of the most common datasets found in the MPMM-related literature (as summarized in \citet{STACHOWIAK2026100905}) and compare them directly to MPMMine.

\emph{CSPLib} \cite{csplib} is a curated collection of combinatorial problems, predominantly formulated as Constraint Programming models \cite{essentialsOfConstraintProgramming}. However, it lacks standardization: directory layouts differ between problems, the nature and quality of supplementary artifacts vary, and most instances provide neither representative solutions nor counterexamples. Unlike MPMMine, using CSPLib entails manually adapting each problem to the target experimental setting, e.g.,\,by rewriting the model into the required form and constructing training examples.

The \emph{MiniZinc challenge} \cite{minizinc2}, while offering well-structured MiniZinc models and instances, is limited to combinatorial problems and was designed primarily for solver evaluation. Unlike MPMMine, it supplies neither solutions/non-solutions nor natural-language descriptions. 

\emph{MIPLIB} \cite{MIPLIB} broadens the scope to continuous and mixed-integer optimization problems; yet, it targets solver performance evaluation and stores models in various low-level formats (MPS, LP) with limited accompanying metadata. Consequently, it suffers from poor standardization and, contrary to MPMMine, it lacks the necessary artifacts for benchmarking MPMM algorithms. 

Other resources face similar limitations: \emph{Netlib} \cite{netlib} provides a~classic but dated collection of Linear Programming (LP) models for solver testing; the \emph{RLFAP} dataset~\cite{rlfap} offers radio link frequency assignment instances in a domain-specific format; and \emph{NL4Opt}~\cite{nl4opt} focuses on extracting optimization formulations from natural-language text but without supplying structured MP models, instances, or solution sets. 

Despite the abundance of solver benchmarks, we identify two datasets closely related to our work:

\emph{PTHG-21}~\cite{simonis_2021_5155465} consists of sample parameters and their corresponding solutions in JSON format for some undisclosed problem formulations. MPMMine is significatly more extensive, as it provides reference models, problem descriptions, and much larger sets of both solutions and non-solutions.

\emph{CP-Bench}~\cite{michailidis_2025_15592407} provides only a single description per problem instance, tying every description to specific parameter values. In contrast, MPMMine reflects a more realistic setting by providing multiple general (parameter-free) descriptions for each problem alongside the reference model, complemented by instance-specific descriptions where appropriate.

\begin{table}
    \centering%
    \caption{Quantitative comparison of datasets commonly used for benchmarking MPMM algorithms.}
    \label{tab:related:datasets}
    \begin{tabular}{l|c|c|c|c|c|c}
        \hline
         Dataset &Problems&Models&Instances&Solutions&Non-solutions&Descriptions  \\\hline
         \textbf{MPMMine} (this work) & 20 & 28 & 387 & 3012625 & 2707553 & 362 \\
         CSPLib                       &  103  &  276  &   103592  &   0  &  0    &  97   \\

         MiniZinc challenge           &  212  &  368  &  1569   &    20  &   0   &  0   \\
         % tutaj niektóre problemy powtarzają się na przestrzeni wielu lat
         %problems = subcatalog count bez duplikatów nazw
         %models = mzn (bez duplikatów (po treści))
         %instances = dzn+json (bez duplikatów (po treści))
         %solutions pliki ".out"
         MIPLIB 2017                   &  461  &  673  &   1529  &   1756   &    0  &   1  \\

         Netlib                       & 60 & 98 & 98  &  0    &   0   &   0  \\ % based on https://www.netlib.org/lp/data/readme   Aggregated the same instance prefixes as variants of the same models
         RLFAP                        & 1  & 30  &  30  &   0  & 0    & 1   \\ % tar pobrany ze strony https://miat.inrae.fr/schiex/rlfap.shtml#inst
         NL4Opt                       &1101&  0  & 0    &  0   &   0  &  1101\\ % https://github.com/nl4opt/nl4opt-competition
         PTHG-21                      & 15 &  0 & 145 &24426 & 41537&  0   \\ % https://doi.org/10.5281/zenodo.5155465
         CP-Bench                     &101 & 101& 101 &  0   &  0   & 101   \\ % https://zenodo.org/records/15592407
         \hline
    \end{tabular}
    
\end{table}

% NL4OPT: Będzie trzeba napisać, że nie ma jednoznacznego wyznacznika, że opisy opisują ten sam problem, więc traktujemy je jako inne, pomimo, ze analiza LLMem pokazuje, że one opisują innymi słowami te same problemy obliczeniowe, ale z innymi wartościami parametrów.

Table \ref{tab:related:datasets} compares the above-mentioned datasets quantitatively with MPMMine based on the contents of their repositories as of 2026-09-01. For CSPLib,\footnote{\url{https://github.com/csplib/csplib}} we count the total number of \texttt{prob*} directories, and files with extensions \texttt{.bprolog}, \texttt{.co}, \texttt{.ecl}, \texttt{.essence}, \texttt{.eprime}, \texttt{.mzn}, \texttt{.minion}, \texttt{.py}, \texttt{.pi}, \texttt{.pl}, \texttt{.mod} in \texttt{models}, and files with extensions \texttt{.params}, \texttt{.dzn}, \texttt{.txt}, \texttt{.dat}, \texttt{.instance} in \texttt{data}, and \texttt{specification.md} files for problems, models, instances, and descriptions, respectively including files compressed in \texttt{.tar.gz}, and \texttt{.zip} archives; it does not contain exemplary solutions and non-solutions. For the MiniZinc challenge,\footnote{\url{https://github.com/MiniZinc/mzn-challenge}} we count the total number of uniquely-named directories, \texttt{.mzn} files, and \texttt{.dzn} and \texttt{.json} files for problems, models, and instances, respectively; it does not provide solutions, non-solutions, and descriptions (except comments in models). For MIPLIB,\footnote{\url{https://miplib.zib.de/}} we count problems as the total number of unique groups if set and records without a group are considered individual problems; models as \texttt{.cpp}, \texttt{.py}, \texttt{.lp}, \texttt{.mod}, \texttt{.ctt}, \texttt{.sas}, \texttt{.mzn}, \texttt{.model}, \texttt{.zpl}, \texttt{.mst}, \texttt{.smp} files in \texttt{models.zip},\footnote{\url{https://miplib.zib.de/downloads/models.zip}} instances as records in collection plus \texttt{.dzn} files in \texttt{models.zip}, solutions as directories in \texttt{solutions.zip}\footnote{\url{https://miplib.zib.de/downloads/solutions.zip}}; we found only a single description file for a problem and no non-solutions. For Netlib,\footnote{\url{https://www.netlib.org/lp/data/readme}}, we count problems as unique text prefixes (without numbers) in the problem name, models and instances as \texttt{.mps} files; it does not hold solutions, non-solutions, and descriptions. RLFAP\footnote{\url{https://miat.inrae.fr/schiex/rlfap.shtml}} consists of just 30 low-level models/instances of a single problem. For NL4Opt,\footnote{\url{https://github.com/nl4opt/nl4opt-competition}} our manual and LLM-based analyzes indicate that 1101 problem descriptions group into tens of types of computational problems differing in narration and parameter values. Due to the lack of an indisputable problem discriminator, we count all of them as unique; however, the actual number of unique computational problems is roughly 10 times smaller. For PTHG-21,\footnote{\url{https://doi.org/10.5281/zenodo.5155465}} we count fields in the JSON files according to the meaning provided in the file format description. CP-Bench\footnote{\url{https://doi.org/10.5281/zenodo.15592407}} consists of 101 text descriptions with reference models.

Existing datasets vary in strengths but each has gaps: broad problem coverage without related artifacts; many low-level models that are hard to read (ambiguous names, no comments, large numeric matrices with unclear meaning); text-to-model benchmarks without systematic model representations and alternatives; and too few solutions for robust benchmarking of example- and query-based MPMM algorithms. Most also lack consistent structure and metadata needed to evaluate MPMM algorithms systematically. MPMMine remedies this with a~purpose-built, consistently structured, and extensible benchmark framework.

\section{MPMMine: standardized benchmarks for MP model mining}\label{sec:mpmm}

\begin{table}[]
    \centering\tabcolsep=0.25em%
    \caption{Problems currently implemented in the MPMMine.}
    \label{tab-problem-list}
    \begin{tabular}{l|p{24em}|l|l|l|l}%{p{4.5em}|p{14em}|p{5em}|p{5.5em}|p{5em}}
        \hline
         Id   & Problem name & Models & Instances & Descriptions & Type \\
         \hline
         P001 & Progressive Party (\citeauthor{csplib:prob013}, \cite{ATHNANORLSOACS}) & 1 & 11 & 19 & Discrete \\
         P002 & Car Sequencing (\citeauthor{csplib:prob001}) & 1 & 110 & 19 & Discrete \\
         P003 & Template Design (\citeauthor{csplib:prob002}) & 2 & 16 & 38 & Discrete \\
         P004 & Low\,Autocorrel.\,Binary\,Seq.\,(\citeauthor{csplib:prob005}) & 2 & 50 & 20 & Discrete \\
         P005 & Golomb Ruler (\citeauthor{csplib:prob006}, \citeauthor{golombrulermzn}, \citeauthor{datagenetics}) & 1 & 10 & 19 & Discrete \\
         P006 & Vessel Loading (\citeauthor{csplib:prob008}) & 1 & 5 & 19 & Discrete \\
         P007 & \raggedright Continuous Knapsack \cite{goodrich2001algorithm} & 1 & 5 & 14 & Continuous \\ 
         P008 & Cutting Stock \cite{gilmore1961linear} & 1 & 5 & 18 & Discrete \\
         P009 & Sphere Packing in a Cube \cite{gensane2004dense} & 1 & 3 & 11 & Continuous \\ 
         P010 & Facility Location \cite{aardal1995capacitated} & 1 & 3 & 11 & Continuous \\ 
         P011 & Schur's Lemma (\citeauthor{csplib:prob015}) & 2 & 45 & 15 & Discrete \\
         P012 & Bus Driver Scheduling (\citeauthor{csplib:prob022}) & 1 & 9 & 19 & Discrete \\
         P013 & Langford's Number (\citeauthor{csplib:prob024}) & 2 & 40 & 37 & Discrete \\
         P014 & Crude Mix \cite{NOH201891} & 2 & 6 & 13 & Continuous \\
         P015 & Feed Blend (original) & 2 & 6 & 32 & Continuous \\
         P016 & Power Management (original) & 1 & 37 & 16 & Continuous \\
         P017 & 3D noughts and crosses \cite{williams2013model,githubGitHubNaraqbwilliamsproblems} & 2 & 12 & 11 & Discrete \\
         P018 & Lost Baggage Distribution \cite{williams2013model,githubGitHubNaraqbwilliamsproblems} & 1 & 3 & 10 & Discrete \\
         P019 & Decentralization \cite{williams2013model,githubGitHubNaraqbwilliamsproblems} & 1 & 3 & 10 & Discrete \\
         P020 & \raggedright Manpower Planning \cite{williams2013model,githubGitHubNaraqbwilliamsproblems} & 2 & 8 & 11 & Discrete \\%
         \hline
         Total & & 28 & 387 & 362 & 6 $\times$ Cont. \\%
         &&&&&14 $\times$ Disc.
    \end{tabular}
\end{table}

The main goal of MPMMine is to offer a standardized collection of benchmarks for diverse set of AI tasks concerned with discovering and maintaining MP models using domain knowledge. Here, domain knowledge denotes any information that specifies the computational problem, including text descriptions, formal documents, equations, symbol sets, example and counterexample solutions, existing MP models, irreducible inconsistent subsystems, and any combination thereof. 
Table~\ref{tab-problem-list} summarizes problems currently included in MPMMine.
Section~\ref{sec:mpmm:rules} introduces the rules of conduct. 
Sections \ref{sec:mpmm:models} and \ref{sec:mpmm:checkers} describe the design of models and checkers. 
Section~\ref{sec:mpmm:artifacts} outlines the procedures for obtaining artifacts.

\subsection{Rules of conduct} \label{sec:mpmm:rules}
We strictly follow the six rules below to build MPMMine.

% Your work should use standard \LaTeX\ sectioning commands:
% \verb|section|, \verb|subsection|, \verb|subsubsection|, and
% \verb|paragraph|. They should be numbered; do not remove the numbering
% from the commands.
% [RS] Dlatego wywaliłem gwiazdki
\subsubsection{Consistency} \label{sec:mpmm:rules:consistency}

The directory structure is consistent across all problems. All problems, models, data, and metadata share the same file structure and file formats.

The dataset is organized within the \texttt{problems} root directory using a four-layer hierarchy, accounting for different modeling approaches and specific data instances as follows:

\begin{enumerate}
\item \textbf{Problem:} Each problem holds a separate directory within the \texttt{problems} directory.
\item \textbf{MP Models:} Each problem contains one or more MP models under the \texttt{models} directory, corresponding to various problem encodings and formulations. Every model is paired with a~checker model providing detailed human-readable diagnostic information about violated and satisfied constraints for a given solution/non-solution.
\item \textbf{Abstract descriptions and instances:} Each model has two subdirectories: \texttt{descriptions} corresponding to instance-independent (i.e.,\,without specific values) problem descriptions, and \texttt{instances} consisting of data files with concrete values for abstract model parameters.
\item \textbf{Artifacts:} This level stores artifacts related to specific instances, including solutions, non-solutions, and instance-specific text descriptions.
\end{enumerate}

The file tree structure is outlined below, with the \texttt{\textbf{[R]}} marker denoting required files:

% Nie wiem czy to nie powinien być zwykły figure
\setlength{\DTbaselineskip}{10pt}
\DTsetlength{0.2em}{1em}{0.2em}{0.4pt}{1.6pt}

\begin{minipage}{0.9\textwidth}%
\footnotesize%
\dirtree{%
.1 problems/.
.2 P000 problem name/.
.3 manifest.json \textbf{[R]}.
.3 references.bib \textbf{[R]}.
.3 models/M000/.
.4 model.mzn \textbf{[R]}.
.4 checker.mzn \textbf{[R]}.
.4 descriptions/.
.5 D000 description.en.md  \textbf{[R]}.
.4 instances/I000 instance name/.
.5 instance.dzn \textbf{[R]}.
.5 descriptions/D000 description.en.md.
.5 solutions/S00000 sol.dzn.
.5 non solutions/N00000 non\_sol.dzn.
}
\end{minipage}

The dataset employs structured, hierarchical identifiers that ensure every file is uniquely referenced. Level-specific ids consist of a single-letter prefix related to file type, and a~three or five-digit unique code, e.g.,\,\texttt{P003} for problem, \texttt{M002} for model, etc. By concatenating the ids at individual levels, a global id is generated, e.g., \linebreak \texttt{MPMMine-P003M002I003S00055} pointing specifically at the 55th solution of the third instance of the second model belonging to the `P003 Template Design' problem. The \texttt{MPMMine-} prefix serves as a~namespace for external referencing. Incomplete ids enable referencing higher levels of the hierarchy.

Each problem includes a \texttt{manifest.json} file providing structured metadata for both human reader and a~machine. The top-level attributes are defined as follows:
\begin{itemize}
    \item \textbf{id} -- Problem id, consistent with the problem directory name,
    \item \textbf{name} -- Descriptive title,
    \item \textbf{tags} -- Problem-specific key-value pairs for indexing; taken from Wikidata \cite{wikidata},
    \item \textbf{features} -- Boolean flags indicating high-level characteristics such as constraints, objective function, whether it is an optimization or satisfiability problem, and the types of variables,
    \item \textbf{alternative\_ids} -- Mapping to related problems in other benchmark datasets,
    \item \textbf{references} -- BibTeX-like paper metadata,
    \item \textbf{links} -- URLs to related resources.
\end{itemize}

The \texttt{references.bib} is a ready-to-use BibTeX definition of the related documents. It consists of the same entries as those included in the \texttt{references} section of \texttt{manifest.json}.

Each \texttt{models} directory contains at least one subdirectory with a~reference MiniZinc model in the \texttt{.mzn} format. These models are instance-independent, defining the problem's logic through abstract parameters rather than hard-coded values.

To support text-to-model benchmarking, every model includes at least one English description in the \texttt{descriptions} subdirectory. These descriptions also use symbols or natural language instead of specific numbers and are tailored to the logic of their corresponding MP model. The descriptions containing a complete mathematical formulation of the problem are explicitly marked in the file name using the \texttt{formal} word.
Descriptions in other languages are optional (indicated by language-code suffixes).

Each MP model includes one or more concrete instances located within the \texttt{instances} directory, following the \texttt{I000 instance\_name} naming convention. The \texttt{instance.dzn} file provides the specific data values required to populate the model's abstract parameters.
While instances are typically feasible, the dataset also includes unsatisfiable instances where no solution exists. These are explicitly flagged by placing the \texttt{\%~UNSATISFIABLE} marker on the very first line of the \texttt{.dzn} file to ensure they are handled correctly during benchmarking.

Each instance is supplemented by domain knowledge artifacts held in the subdirectories:
\begin{itemize}
    \item \texttt{descriptions} -- The instance-level descriptions with symbols set to concrete values.
    \item \texttt{solutions} -- The unique solutions (e.g.,\,\texttt{S00000 sol.dzn}); it targets 10,000 solutions per instance; though fewer may be provided for smaller solution spaces or computationally difficult instances; %No solutions may be provided for instances computationally hard yet not proven to be unsatisfiable. 
    this directory is omitted for unsatisfiable instances.
    \item \texttt{non solutions} -- The unique non-solutions (e.g.,\,\texttt{N00000 non\_sol.dzn}); the default count is 10,000, but the directory is excluded for unconstrained problems.
\end{itemize}
Consult Section \ref{sec:mpmm:artifacts} for details on how these files were created.

\subsubsection{Standardization}\label{sec:mpmm:rules:standardization}

The dataset relies exclusively on open standards to ensure interoperability and accessibility: MiniZinc \cite{minizinc1} for MP models, CommonMark \cite{commonmark} for text descriptions, JSON \cite{json} for metadata, Wikidata \cite{wikidata} for tag taxonomy, ISO-639-1 \cite{iso639-1} for two-letter language codes, ISO-8601 \cite{iso8601} for date and time format, and American English for all files with unspecified language. 

\subsubsection{Completeness}\label{sec:mpmm:rules:completeness}

All data are complete and accurate. Partial problem descriptions, missing parts of MP models, missing parameter values, and NA values in solutions are strictly prohibited.

\subsubsection{Extensibility}\label{sec:mpmm:rules:extensibility}

The directory structure and file formats are chosen to facilitate extensibility with new problem types, new MP models for existing problems, new instances for existing MP models, new domain knowledge artifacts, and new types of artifacts.

\subsubsection{Open Process}\label{sec:mpmm:rules:open}

Updates, fixes, and extensions to the dataset are open to the community through pull~requests.

\subsubsection{Version control and backward compatibility}\label{sec:mpmm:rules:vcs}

To ensure experimental reproducibility and facilitate result comparisons, the dataset is version-controlled via Git \cite{git}. Major milestones are identified using tags. Once a~file is associated with a tagged release, it is subject to strict immutability rules to ensure~backward~compatibility:
\begin{itemize}
    \item \textbf{Models} -- Modifying the code of a tagged model is prohibited, though formatting and comments may be updated; to correct a bug, create a new derived model with a unique~id.
    \item \textbf{Instances} -- Parameter values in tagged instances are permanent; if a bug is identified, create a new instance with a new id; formatting adjustments remain permissible.
    \item \textbf{Descriptions} -- Tagged textual descriptions are fixed and cannot be altered.
    \item \textbf{Solutions/Non-solutions} -- Data values within these files are immutable; to address an erroneous entry, the file must be prepended with a \texttt{\% ERROR: <description>} comment; a~corrected version can then be issued as a new file.
\end{itemize}

\subsection{Models}\label{sec:mpmm:models}

MPMMine consists of 28 MiniZinc models out of which 8 models use global constraints, 10 models include symmetry breaking constraints, and 3 models guide the optimization with implied constraints. Models involve continuous and discrete variables, where a few actually involve variables of different types. This shows that MPMMine models cover a~wide variety of MP classes, including CP and MILP/LP. The MiniZinc tool offers seamless conversion capabilities between model classes. Although the size of an MP model is typically limited to a dozen symbols (variables and parameters) and a dozen constraints, the data provided in the corresponding instances expands models to sizes varying from tens to even hundreds of thousands of variables and low-level constraints due to the common use of quantifiers (e.g.,\,\texttt{forall}), set algebra (e.g.,\,Cartesian product), and global constraints. % Define size of a model?

\subsection{Model checkers}\label{sec:mpmm:checkers}

Every model is paired with a MiniZinc model checker which when provided with a corresponding instance and (non-)solution can assess its (in-)feasibility. Model checkers enable also optional validation of the objective function value. Checkers return a control string \texttt{CORRECT: all constraints hold.} when there are no violations of constraints by a solution, and a~control string \texttt{INCORRECT: See error log above for specific violations.} when a solution violates at least one constraint. The output stream holds a human-readable list of all detected violations, e.g.,\,checking \texttt{MPMMine-P001M001I009N00002} yields output:

\begin{verbatim}
ERROR: visit[1,3] (10) is not in hosts.
ERROR: Host logic failed for boat 1 at time 1
ERROR: Host logic failed for boat 1 at time 3
ERROR: Boats 1 and 9 met too many times. Mutual places: [(1, 6), (2, 1)]
ERROR: Symmetry breaking failed between Period 1 and Period 2.
  Mismatch found at Boat index: [1, 4, 5, 7, 9, 10, 11]
  Period 1: [6, 2, 3, 1, 2, 6, 12, 8, 6, 3, 8, 12]
  Period 2: [1, 2, 3, 2, 8, 6, 3, 8, 1, 12, 6, 12]
ERROR: Symmetry breaking failed between Period 3 and Period 4.
  Mismatch found at Boat index: [1, 4, 5, 7, 9, 10, 11]
  Period 3: [10, 2, 3, 3, 1, 6, 2, 8, 8, 6, 2, 12]
  Period 4: [1, 2, 3, 6, 12, 6, 8, 8, 2, 1, 1, 12]
INCORRECT: See error log above for specific violations.
\end{verbatim}

\subsection{Domain knowledge artifacts}\label{sec:mpmm:artifacts}

MPMMine provides various types of domain knowledge artifacts: exemplary solutions and non-solutions for the example-driven discovery and maintenance of MP models, as well as English problem descriptions for benchmarking text-to-model scenarios.

% Kolmogonorov-Smirnov test is for continuous distributions only. When plotting the distributions of particulal variables, the constraints make it difficult for the discrete values to be uniform
% \begin{figure}
% \begin{subfigure}[t]{0.245\textwidth}
% \centering\includegraphics[width=\textwidth]{img-P006I002-int-sol-40var.png}
% \caption{P006M001I002\\solutions}
% \label{fig-exemplary-solutions-006I002-int-sol-40var}
% \end{subfigure}
% \begin{subfigure}[t]{0.245\textwidth}
% \centering\includegraphics[width=\textwidth]{img-P006I002-int-non-sol-40var.png}
% \caption{P006M001I002\\non-solutions}
% \label{fig-exemplary-solutions-P006I002-int-non-sol-40var}
% \end{subfigure}
% \begin{subfigure}[t]{0.245\textwidth}
% \centering\includegraphics[width=\textwidth]{img-P007I002-cont-sol-10var.png}
% \caption{P007M001I002\\solutions}
% \label{fig-exemplary-solutions-007I002-cont-sol-10var}
% \end{subfigure}
% \begin{subfigure}[t]{0.245\textwidth}
% \centering\includegraphics[width=\textwidth]{img-P007I002-cont-non-sol-10var.png}
% \caption{P007M001I002\\non\nobreakdash-solutions}
% \label{fig-exemplary-solutions-P007I002-cont-non-sol-10var}

% \end{subfigure}
% \caption{4D PCA landscape of solutions and non-solutions of problems P006 and P007; color corresponds to the fourth dimension.}
% \label{fig-exemplary-solutions}
% \end{figure}

\subsubsection{Exemplary solutions}\label{sec:mpmm:artifacts:sol}

Unique exemplary solutions are sampled independently for each MP model instance from the feasible region of the model given the parameter values from the instance. Since strict uniform sampling in highly-constrained high-dimensional spaces typically fails using common techniques like Monte Carlo \cite{robert2004monte}, Hit-and-Run \cite{Smith1996}, and Gibbs \cite{Lee2021}, we employ the best-effort uniform sampling technique described below. 

We first encode the model using Linear Programming (LP) or Mixed-Integer LP (MILP), depending on the variable domains. We mark variables as \emph{output}, i.e.,\,directly corresponding to the values of variables in the original model, and \emph{non-output}, i.e.,\,helper variables introduced through transformation. Then, we collect feasible solutions of this model by tracking intermediate solutions found by the solving algorithms Branch-and-Cut \cite{Mitchell2002} and Simplex \cite{Dantzig1963} in the Gurobi solver \cite{gurobi}. To avoid bias in distribution, we draw a~random objective function every 50 (MILP) or 1 (LP, problems: P018, P019) distinct solutions found. Two solutions are considered distinct if the values of any \emph{output} variable differ by at least $10^{-6}$.  For MILP models, this procedure collects solutions at some distance from the constraints. For LP models, this collects the vertices of the simplex of constraints. In LP models, to obtain exemplary solutions from the interior of the simplex, we apply postprocessing: once fifty duplicate solutions have been encountered in a~row, a~random subset of known solutions is drawn with replacement, and a~random convex combination of these selected solutions is then constructed and stored.
Solutions near the feasibility boundary are prone to numerical instability, where tiny calculation errors may alter feasibility. To avoid this, we tighten all constraints in LP models, shifting their right-hand sides by $10^{-4}$ in the appropriate direction.

% To wyleci jak się zmieni sposób walidacji dystrybucji
%\emph{DROP THIS:}
%We visually validate the uniformity of the obtained sample by reducing it to 4 dimensions using Principal Component Analysis (PCA). Figure~\ref{fig-exemplary-solutions-006I002-int-sol-40var} shows a~solution distribution for a~representative model with 40 integer-only variables. Although the visualization explains only $36\%$ of the variance, the cloud of points uniformly fills the entire feasible region without oversampling or undersampling some locations. 
%Figure \ref{fig-exemplary-solutions-007I002-cont-sol-10var} shows an analogous visualization for a continuous model with 10 variables. It explains $52\%$ of the variance. Here, the examples also look uniformly distributed, and the point cloud in the middle consists of the convex linear combinations of solutions.

\begin{table}
    \centering%
    \caption{The test for uniformity of solution sampling procedure; sample and reference set sizes, the p-value of the energy test for equality of distributions, the p-value corrected using Holm-Bonferroni method, test conclusion, the goodness-of-fit index. }
    \label{tab:mpmm:artifacts:sol:test}
    \begin{tabular}{l|r|r|r|r|c|r}
        \hline
         Instance     & Sample size & Reference set size & p-value & Corrected p-value & Uniform & GIF\; \\\hline
         P001M001I006 &      10,000 &           100,000 & 0.000 & 0.000 & False & 0.997 \\
         P002M001I001 &      10,000 &           100,000 & 0.000 & 0.000 & False & 0.996 \\
         P003M002I001 &      10,000 &           100,000 & 0.000 & 0.000 & False & 0.910 \\
         P004M001I008 &      10,000 &           100,000 & 0.000 & 0.000 & False & 0.995 \\
         P005M001I006 &      10,000 &           25,645 & 0.004 & 0.027 & True & 0.998 \\
         P006M001I003 &      10,000 &           100,000 & 0.001 & 0.007 & True & 0.997 \\
         P007M001I001 &      10,000 &           100,000 & 0.000 & 0.000 & False & 0.922 \\
         P008M001I001 &      10,000 &           100,000 & 0.000 & 0.000 & False & 0.996 \\
         P009M001I001 &      10,000 &           100,000 & 0.692 & 1.000 & True & 0.999 \\
         P010M001I001 &      10,000 &           100,000 & 0.116 & 0.698 & True & 0.999 \\
         P011M001I003 &      10,000 &           100,000 & 0.003 & 0.020 & True & 0.997 \\
         P012M001I007 &      10,000 &           27,349 & 0.000 & 0.000 & False & 0.999 \\
         P013M001I005 &      10,000 &           17,792 & 0.000 & 0.002 & True & 0.999 \\
         P014M001I001 &      10,000 &           100,000 & 0.000 & 0.000 & False & 0.990 \\
         P015M001I001 &      10,000 &           100,000 & 0.000 & 0.000 & False & 0.970 \\
         P016M001I001 &      10,000 &           100,000 & 0.196 & 0.982 & True & 0.999 \\
         P017M001I001 &      10,000 &           100,000 & 0.313 & 1.000 & True & 0.999 \\
         P018M001I002 &      10,000 &           100,000 & 0.433 & 1.000 & True & 0.999 \\
         P019M001I002 &      10,000 &           24,989 & 0.610 & 1.000 & True & 1.000 \\
         P020M001I001 &      10,000 &           100,000 & 0.000 & 0.000 & False & 0.983 \\
         \hline
    \end{tabular}
    
\end{table}

We assess sampling uniformity using the energy test for equality of high-dimensional distributions \cite{szekely2004testing,EnergyDistance2016} on representative instances (one per problem) whose feasible regions can be efficiently enumerated. We build an independent reference set of feasible solutions, capped at 100{,}000 to limit computational cost. The null hypothesis is that sample and reference solutions come from the same distribution. We use a~family-wise significance level of $\alpha=0.001$ and apply the Holm-Bonferroni correction \cite{HolmCorrection} for multiple tests. Because the required distance matrix scales quadratically, we subsample 5,000 solutions and estimate significance with 10,000 permutations. 
Following the recommendations of \citet{Benjamin2017,Lin2013}, we choose a very small $\alpha$, since with large samples statistical tests can reject the null hypothesis because of tiny numerical noise that has no practical engineering relevance. In our setting, this noise is evidently nonzero: the energy distance (part of the test statistic) sums pairwise distances over thousands of variables, so small numerical errors in individual dimensions can accumulate. 
Such errors can result from different pseudorandom seeds, integer quantization, constraint shifts, and solver tolerances. \citet[Sec.~5]{10.1145/3205455.3205480} show that even an arbitrarily thin but non-empty margin around a constraint has a volume that grows exponentially and diverges as the number of variables increases. Thus, shifting a constraint by an arbitrarily small amount (as stated above) can remove many feasible solutions and change the induced distribution enough to falsely reject uniformity.
Those same references also recommend complementing $p$-values with an effect-size metric; accordingly, we additionally report the goodness-of-fit (GIF) index (0 = worst, 1 = best) to quantify the level of agreement.

Table \ref{tab:mpmm:artifacts:sol:test} summarizes the instances along with the gathered statistics. Ten out of the twenty tests indicate uniform sampling. Nevertheless, since GIF achieves values between 0.91 and 1.0, reflecting a strong agreement with the uniform distribution, we attribute the unsuccessful tests to the phenomena described above. Therefore, we conclude that these results provide sufficient evidence that the solution instances in MPMMine are approximately uniformly distributed.

The large sizes of solution sets in MPMMine further permit flexible tailoring of the empirical distribution to specific experimental requirements by applying biased sampling procedures over the solutions in the dataset. This capability may be particularly valuable for use cases such as evaluating an algorithm's robustness under biased, imbalanced, incomplete, and/or noisy data conditions.

\subsubsection{Exemplary non-solutions}\label{sec:mpmm:artifacts:nonsol}

MPMMine provides exemplary non-solutions that closely resemble valid solutions but contain small, deliberate errors that violate constraints; variable values never violate their domains. This ensures that the non-solutions are specific to the given model and more clearly highlight the locations of violated constraints.
For integer-only problems, exemplary non-solutions are obtained by randomly selecting a~small subset of variables in a~valid solution and resampling their values from the corresponding domains, while guaranteeing that the resulting assignment is both unique and infeasible. This results in non-solutions that are not obviously defective, even for a human expert. This way, the non-solutions are highly informative for constraint learning, in contrast to random non-solutions that often lie much farther from the feasible region, and the errors in them are straightforward to identify by a~human expert.
For continuous/mixed problems, exemplary non-solutions are created by sampling random values for all discrete and continuous variables from their respective domains, again ensuring uniqueness and infeasibility.
To handle possible numerical issues caused by LP models' non-solutions lying near constraints, the sampling is performed with constraints loosened by adjusting their right-hand sides by $10^{-4}$, or by making a small ribbon $2\cdot10^{-4}$ width when relaxing an equality constraint.

\subsubsection{Additional notes on the (non-)solution generation}
The choice of the Gurobi solver for sampling solutions/non-solutions is not dictated by the limitation to LP/MILP. Actually, the MiniZinc models are more robust than LP/MILP, however, MiniZinc provides a transcoder to LP/MILP that Gurobi may handle. This way, we may efficiently handle problems involving continuous variables, as a more traditional approach in CA involving CP solvers, like Gecode, often fails in the presence of continuous variables.

Problem P004 is effectively unconstrained: Its constraints serve solely to determine the objective value, making P004 suitable for benchmarking objective-related~tasks. Due to the above reasons instead of sampling, we took 10{,}000 best solutions according to the objective function.

%Non-solutions for P012M001I005 are calculated using the procedure for solutions with an altered constraint 
%\begin{minted}{text}
%constraint forall(j in 0..num_work-1) (
%   sum(i in 1..num_shifts) (x[i]*bool2int(j in shifts[i])) != 1 % equality changed to inequality
%);
%\end{minted}

Each solution and non-solution includes a commented objective function value (if any), calculated from the variable values of the associated solution or non-solution.

\subsubsection{Natural-text descriptions}\label{sec:mpmm:artifacts:desc}

The MP models are accompanied by text problem descriptions for text-to-model mining tasks. All descriptions are curated and validated by humans for consistency with their respective MP models. Different texts are provided to handle different levels of abstraction (general vs instance-specific), different levels of formalism (natural language vs formal natural language), and different narration styles (created by different experts and/or LLMs). These differences contribute to overall diversity and may reveal that performance of a particular text-to-model approach depends on specific types of input data.

At least one description was produced for each MP model. Some were authored by human experts---either taken from the original repository when the MP model was reused, or newly written by the MPMMine developers. Additional descriptions were generated with several Large Language Models (LLMs): llama3.3 \cite{grattafiori2024llama3herdmodels}, deepseek-r1 \cite{Guo2025}, gemma3 \cite{gemmateam2025gemma3technicalreport}, gemma4:26b \cite{gemmateam2026gemma4technicalreport}, qwen3.8:27b \cite{qwen38}, gpt-oss \cite{openai2025gptoss120bgptoss20bmodel}, nemotron-3-nano \cite{nvidia2025nemotron3nanoopen}, nemotron3:33b \cite{nvidia2026nemotron3nanoomni}, and mistral-small3.2 \cite{MistralAI2025Small32}, which were given the handcrafted description and the reference MP model as input. All LLM outputs were manually revised by a~human expert to ensure consistency with the MP model and to eliminate incorrect or undesired content. Every LLM was queried with the same prompts, detailed in Appendix~\ref{app:llm}. The use of multiple LLMs resulted in heterogeneous narrative styles and degrees of formality.

\subsection{Distribution package}

%introduce sqlite+zst; introduce mpmmine-py
For use in the development, verification, and benchmarking of MPMM algorithms, we recommend using the compressed SQLite-backed package\footnote{\url{https://github.com/MPMMine/MPMMine/releases}} alongside the \texttt{mpmmine-py}\footnote{\url{https://github.com/MPMMine/mpmmine-py}} Python package for data access. This setup offers lower latency than direct file-system access while providing a highly compressed representation. In contrast, bare repository clones consume roughly 5-6x more disk space and result in significantly higher latency. Consult the \texttt{mpmmine-py} documentation for usage instructions.

\section{Experimental validation of the MPMMine benchmark dataset}\label{sec:leaderboard}

To demonstrate the benchmarking capabilities of the MPMMine dataset, we evaluate several constraint acquisition algorithms on it.

\subsection{The challenge of evaluation MPMM methods}

There is no consensus on the best algorithm to discover MP models and even the most effective conformance checking approach. Researchers use many bespoke benchmarks with limited evidence that they are unbiased or not tailored to specific algorithms. The community often uses custom model representations. Implementations are rarely released as open, durable artifacts, often becoming unavailable within a~few years, and are written in heterogeneous programming languages with inconsistent I/O interfaces, hindering rigorous comparisons~\cite{STACHOWIAK2026100905}.

\begin{figure}
    \centering
    \includegraphics[width=1.0\linewidth]{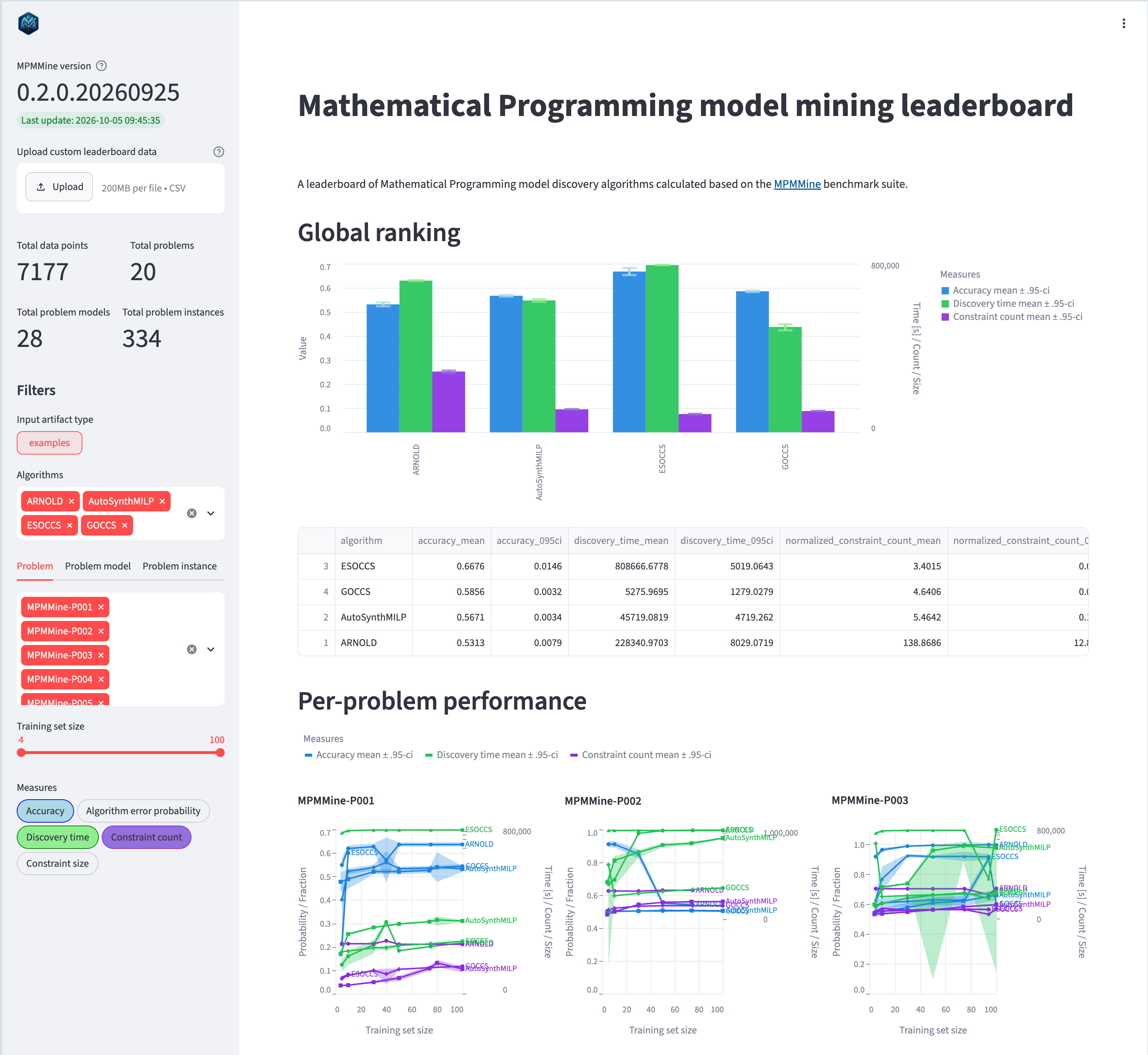}
    \caption{MPMMine Leaderboard, an interactive dashboard for the evaluation of MP model discovery algorithms.}
    \label{fig:leaderboard:demo}
\end{figure}

In this work, we propose unifying experimental protocols using MPMMine. We use the MPMMine Leaderboard \cite{MPMMineLeaderboard}, a~software framework and experimental-results repository designed to enable rigorous and equitable benchmarking of existing and forthcoming methods for MP model discovery.\footnote{\url{https://github.com/MPMMine/Leaderboard}}$^,$\footnote{\url{https://mpmmine.streamlit.app}} Figure \ref{fig:leaderboard:demo} shows the interactive dashboard, which provides direct access to the results of MPMM discovery algorithms and supports multidimensional analyzes, including (but not limited to) evaluation metrics such as accuracy, error probability, and discovery time, as well as scalability with respect to training-set size and numbers of variables and parameters. In addition, it facilitates configurable cross-method comparisons and offers access to the resulting MP models.

\subsection{Evaluation methodology}

MPMMine Leaderboard is built on several principles that contribute to reproducibility, standardization, longevity, openness, and unification of the results. 

\subsubsection{Reproducibility}

To achieve reproducibility, we divide solution sets from MPMMine into training and test splits within a $k$-fold cross validation loop with a fixed random seed. This way, although the data split is random, it is fully reproducible, and all algorithms are trained and tested using the same pieces of data. $k=5$ to handle computational cost.

\subsubsection{Standardization}
Since the reference implementations use different programming languages, technologies, software libraries, versions, input and output formats, etc., we unify the algorithmic interface by introducing a layer of algorithm adapters. The Leaderboard README provides all the necessary instructions for adapter preparation and extension of the Leaderboard with a new algorithm.

We unify the data format available to each algorithm to prevent information loses or gains due to changes in data structures, etc., making the comparison fair. In particular, the algorithm adapters are provided with training artifacts such as (1) MiniZinc data files (\texttt{.dzn}) for parameter values of a problem instance, exemplary solutions, and exemplary non-solutions; and (2) Problem descriptions in CommonMark.

The adapter is free to use any part of the artifacts provided to build the input for the target algorithm and is required to convert algorithm's output into an MP model in the MiniZinc model format \texttt{.mzn}.

The test artifacts consist of the same type of data as training artifacts plus the reference MP model in the MiniZinc model format \texttt{.mzn}. The algorithm adapters do not have access to the test data. Testing is done solely by the Leaderboard evaluator.

\subsubsection{Longevity}
Algorithms are run within an isolated environment of a Docker container to reduce potential inference from other software, libraries, services, and system configuration. The contents of the Docker container is algorithm-specific to satisfy all requirements of the implementation, while minimizing inference from other factors. The images are prepared based on open-source software if applicable, e.g., linux as a containerized OS, equipped with packages required by the implementation.

Open source code plus standardized environment guarantee longevity of the algorithmic setup while enabling adaptation to changes in hardware, e.g., CPU architecture. As source code is compiled (if applicable) on the target platform, it uses the machine code of this platform. On the other side, frozen contents of the Docker images prevent side effects in algorithms' behavior caused by changes in software.

\subsubsection{Open Process}
The Leaderboard is based on reference implementations of algorithms attached as supplementary material to research papers. All included implementations are open source. They are referenced in the Docker image using either a~fork of the original GitHub repository or a newly created repository within the MPMMine organization if the original source code was not published via GitHub. We do not include closed-source binaries of the reference algorithm implementations, as their contents are difficult to verify.

We avoid modifying the reference implementations. Modifications to legacy code are only allowed if they are required to compile or run the code within a Docker container, and they do not change the algorithm behavior.

\subsection{Leaderboard results}

\begin{table}
    \centering%
    \caption{The performance measures for the selected algorithms, calculated using problems P001-P020 in MPMMine.}
    \label{tab:leaderboard:results}%
    \begin{tabular}{l|r|r|r|r}
         Algorithm & Accuracy (rank) & Error prob. (rank) & Discovery time [s] (rank) & Constraint count (rank) \\\hline
         ARNOLD    &    0.531 (4)    & 0.245       (3)    & 227344             (3)    & 174.35           (4) \\
         AutoSynthMILP& 0.596 (2)    & 0.048       (1)    &  48145             (2)    &   4.97           (3) \\
         ESOCCS    &    0.668 (1)    & 0.910       (4)    & 808667             (4)    &   3.40           (1) \\
         GOCCS     &    0.586 (3)    & 0.006       (2)    &   5276             (1)    &   4.64           (2) \\
    \end{tabular}
    
\end{table}

We assessed four algorithms: ARNOLD \cite{ijcai2019p158}, AutoSynthMILP \cite{PAWLAK20171141}, ESOCCS \cite{pawlak2019synthesis,10.1145/3205455.3205504}, and GOCCS \cite{pawlak2018synthesis}, across all models and all instances for problems P001–P020 in MPMMine. Each algorithm is supplied with training sets containing example solutions and, where relevant, non-solutions, with sizes ranging from 4 to 100. Table~\ref{tab:leaderboard:results} reports overall aggregates for test-set accuracy, the probability of runtime errors, discovery time, the number of constraints found, and the text size of the discovered MP models after conversion to MiniZinc and normalization; see \cite{MPMMineLeaderboard} for definitions of these metrics.
The Leaderboard provides finer-grained results via charts and accompanying tables for individual problems, models, instances, and specific training-set sizes. We do not discuss comparative algorithm performance here, as it lies outside the scope of this paper.

% insert leaderboard here

%% Nazwa algorytmu, cytowanie w formacie authoryear, link do repozytorium?, język źródłowy programowania(żeby pokazać różnorodność), accuracy, error probability, czas wykonania algorytmu, komentarz (np że ten Arnold działa tylko na INT problemach

\section{Discussions and limitations}

MPMMine comprises 20 problems, which by no means span the full range of known computational problems. This initial collection was deliberately selected to reflect problems frequently studied in the MPMM literature (see the surveys by \citet{STACHOWIAK2026100905,Cappart2025}), include several new problems, and ensure diversity. Notably, six problems are continuous, distinguishing MPMMine from other benchmarks employed in MPMM research. Our long-term objective is to scale MPMMine to encompass at least 50 problems and several MP models per problem. To this end, we will continue advancing MPMMine internally and actively invite community contributions of new artifacts via pull requests.

% Uniformity
MPMMine has limitations. The sampling procedures based on random objective functions and convex combinations for continuous cases do not guaranty perfectly uniform distributions, although our statistical analysis indicates that they are approximately uniform. We also tested several \emph{standard} uniform sampling methods that were unsuccessful: Monte Carlo \cite{robert2004monte}, which failed to find feasible solutions for MP models with more than 6–7 variables; a~modified Hit-and-Run algorithm \cite{Smith1996} adapted to integer variables, which yielded biased distributions on discrete instances; and Gibbs sampling \cite{Lee2021}, which was computationally prohibitive.

The results of IEEE-754 floating‑point computations are sensitive to several factors, e.g.,\,the order of operations. To mitigate numerical instability in sampled solutions and non‑solutions for continuous problems, we avoided sampling examples too close to the constraints. This mean was sufficient to ensure robust classification of the examples based on the constraints in the reference MP model, even under varying computation orders. For discrete problems, we did not use constraint tightening or relaxation, as it frequently prevented us from producing discrete solutions that lie on the constraints. Nonetheless, discrete models can also be affected by floating‑point inaccuracies when constraints contain non‑discrete parameters. Note that the correctness of the classification of the sampled solutions and non-solutions was verified using the corresponding checker models.

MiniZinc \cite{minizinc1} was adopted as the reference modeling format because of its expressive high-level constructs, extensive support for model translation, and its widespread use in the MPMM literature \cite{STACHOWIAK2026100905}. However, committing to a~single language can hinder uptake in communities that rely on other ecosystems such as AMPL \cite{fourer2003ampl}, GAMS \cite{gams}, ZIMPL~\cite{zimpl}, and ESSENCE~\cite{essence}, illustrating a~trade-off between consistency and completeness. For the simplicity of maintaining MPMMine, we ultimately chose to restrict model formats to MiniZinc only.

Although the current problem descriptions support benchmarking for various text-to-model and mixed-artifact-to-model settings, this component likely requires further augmentation with annotated named entities to be practically applicable and to satisfy the demands of text-to-model methods, including Ner4Opt \cite{ner4opt}, NL4Opt \cite{nl4opt}, and OptiMUS \cite{pmlr-v235-ahmaditeshnizi24a}.

Currently, MPMMine directly enables benchmarking passive MPMM algorithms, passive text-to-model algorithms, but also example-driven comparison of MP models. For active MPMM setting, it provides the reference MP models for the use by the query verification oracle, and simplifies finding arguments by directly providing exemplary solutions and non-solutions. It supports a~classic two-class approach, where solutions and non-solutions are known, a one-class approach, where only solutions are known, and an unlabeled approach, where one needs to combine and shuffle solutions with non-solutions in the desired proportions to obtain the training and test sets.

\section{Conclusions and future work}\label{sec:conclusions}

We identified deficiencies in benchmarks commonly employed in MP model mining research, such as ad-hoc benchmarks, custom model representations, and undocumented experimental protocols. 
Based on these insights, we developed \emph{MPMMine} explicitly for evaluating MP model mining methods, following stringent quality criteria that ensure a~uniform file structure, standardized formats, extensibility, openness, and backward~compatibility.
It stands out from existing benchmarks through several distinctive features. 

Encoding MP models in a high-level modeling language exposes structural information on parameter-variable relationships that is typically obscured in low-level formulations, thereby providing a~substantive basis for verifying selected properties of the MPMM algorithm. Using multiple MP encodings of the same underlying problem further facilitates the evaluation of MPMM performance under varying representations. In addition, parameterized MP models combined with multiple textual problem descriptions enable systematic assessment of how algorithmic performance scales with the size and heterogeneity of the training corpus. Although increased training volume commonly improves model fit, constructing MP models that remain valid across diverse parameterizations is non-trivial.

Algorithmic performance is also strongly affected by the distribution of solutions and non-solutions: samples concentrated in a~narrow region of the feasible set may induce an unintended overemphasis on that region during learning. While deliberate distribution shaping supports the study of different learning regimes, directly sampling from the high-dimensional, heavily constrained solution space of an MP model is technically difficult. To mitigate this, a~large, approximately uniform, precomputed sample provided by MPMMine can serve as a~base for subsequent distribution shaping (e.g., via oversampling). This led to substantial dataset expansion, reaching approximately six million files and a~total size of 20 GB. Packaging the dataset as a~single compressed archive reduces access latency several times relative to direct file-system access and decreases disk usage by a factor of approximately 5-6. The \texttt{mpmmine-py} package provides transparent access to individual artifacts irrespective of the underlying dataset packaging.

MPMMine was designed with extensibility in mind. Beyond its initial 20 problems, we plan to broaden coverage to additional domains, including computational physics, resource allocation, network flow, and encryption, and to offer larger instances to assess algorithmic scalability with respect to instance size. We will also release a~dedicated software suite to support the creation of subsequent MPMMine versions and to enable systematic, comparative evaluation of new MP model mining algorithms within a unified experimental framework.

\begin{acks}
This research was funded in whole or in part by the National Science Centre, Poland, grant number \linebreak 2023/50/E/ST6/00237. For the purpose of Open Access, the author has applied a~CC-BY public copyright licence to any Author Accepted Manuscript (AAM) version arising from this submission.
\end{acks}

\section*{CRediT authorship contribution statement}

\textbf{Tomasz P.~Pawlak}: Conceptualization, Funding acquisition, Investigation, Methodology, Resources, Supervision, Validation, Writing -- original draft.
\textbf{Rafał Stachowiak}: Data curation, Investigation, Software, Visualization, Writing -- original draft.

\section*{Declaration of Generative AI and AI-assisted technologies in the writing process}
During the preparation of this work the authors used a large-language-model writing assistant to edit prose, correct grammar and spelling, and check formatting. The authors have reviewed and edited all AI-generated content as needed and take full responsibility for the content of the publication.

\section*{Declaration of competing interest}
The authors declare the following financial interests/personal relationships which may be considered as potential competing interests:

Tomasz P. Pawlak and Rafał Stachowiak report that financial support was provided by the National Research Centre, Poland.

\printbibliography

\appendix

\section{LLM prompts for description generation}\label{app:llm}

Below, we present the prompts used to query LLMs for the automatic component of problem description generation, as detailed in Section \ref{sec:mpmm:artifacts:desc}. The full system prompt is shown, and the user prompt corresponds to its last five lines.
\begin{verbatim}
Given the text description of a computational problem in Markdown and the 
reference Mathematical Programming model in MiniZinc, create an alternative 
text description of this problem in Markdown that includes all details from 
the original description and the MiniZinc model but use different words. Do 
not copy the MiniZinc code. Do not introduce numbers from the MiniZinc model; 
instead use symbols for entities extracted from the MiniZinc model. Keep the 
output text of similar length as the original text description. Respond with 
the output text description only.
The data format is as follows:
Description:
<original description here>
==========
MiniZinc model:
<minizinc model here>
\end{verbatim}

\end{document}